\documentclass[conference]{ieeeconf}
\IEEEoverridecommandlockouts
\usepackage{cite}
\usepackage{amsmath,amssymb,amsfonts}
\usepackage{algorithmic}
\usepackage{graphicx}
\usepackage{textcomp}
\usepackage{xcolor}
\def\BibTeX{{\rm B\kern-.05em{\sc i\kern-.025em b}\kern-.08em
    T\kern-.1667em\lower.7ex\hbox{E}\kern-.125emX}}
\graphicspath{{./figures/}}
\usepackage{glossaries}
\usepackage{subcaption}

\usepackage{tikz}
\usepackage{textcomp}
\usepackage{hyperref}
\usepackage{lipsum}

\newcommand\copyrighttext{%
  \footnotesize \textcopyright 2023 IEEE. Personal use of this material is permitted. Permission from IEEE must be obtained for all other uses, in any current or future media, including reprinting/republishing this material for advertising or promotional purposes, creating new collective works, for resale or redistribution to servers or lists, or reuse of any copyrighted component of this work in other works.
  DOI: 10.1109/ITSC57777.2023.10422421}
\newcommand\copyrightnotice{%
\begin{tikzpicture}[remember picture,overlay]
\node[anchor=south,yshift=10pt] at (current page.south) {\fbox{\parbox{\dimexpr\textwidth-\fboxsep-\fboxrule\relax}{\copyrighttext}}};
\end{tikzpicture}%
}

\usepackage{tikz,xcolor,hyperref}
\definecolor{lime}{HTML}{A6CE39}
\DeclareRobustCommand{\orcidicon}{
	\begin{tikzpicture}
	\draw[lime, fill=lime] (0,0) 
	circle [radius=0.16] 
	node[white] {{\fontfamily{qag}\selectfont \tiny ID}};
	\draw[white, fill=white] (-0.0625,0.095) 
	circle [radius=0.007];
	\end{tikzpicture}
	\hspace{-2mm}
}
\foreach \x in {A, ..., Z}{\expandafter\xdef\csname orcid\x\endcsname{\noexpand\href{https://orcid.org/\csname orcidauthor\x\endcsname}
			{\noexpand\orcidicon}}
}

\newacronym{ADS}{ADS}{Automated Driving System}
\newacronym{ODD}{ODD}{Operational Design Domain}
\newacronym{VRU}{VRU}{vulnerable road user}
\newacronym{VRUs}{VRU}{vulnerable road users}
\newacronym{ADAS}{ADAS}{Advanced Driving Assistant Systems}
\newacronym{LKA}{LKA}{Lane Keeping Assistance}
\newacronym{LDW}{LDW}{Lane departure warning}
\newacronym{LCA}{LCA}{Lane Change Assistant}
\newacronym{ADSs}{ADS}{Automated Driving Systems}
\newacronym{SAE}{SAE}{Society of Automotive Engineers}
\newacronym{DDT}{DDT}{Dynamic Driving Task}
\newacronym{NHTSA}{NHTSA}{National Highway Traffic Safety Administration}
\makeatletter 
\newcommand{\linebreakand}{%
  \end{@IEEEauthorhalign}
  \hfill\mbox{}\par
  \mbox{}\hfill\begin{@IEEEauthorhalign}
}
\makeatother 

\author{Novel Certad\orcidN{} \emph{Graduate Student Member, IEEE}, Sebastian Tschernuth\orcidS{} \emph{Student Member, IEEE}, \\and Cristina Olaverri-Monreal\orcidC{} \emph{Senior Member, IEEE}%
\thanks{Chair Sustainable Transport Logistics 4.0, Johannes Kepler University Linz, Altenberger Straße 69, 4040 Linz, Austria.
\texttt{\{novel.certad\_hernandez, sebastian.tschernuth, cristina.olaverri-monreal\}@jku.at}}%
}

\title{\LARGE \bf
Extraction of Road Users' Behavior From Realistic Data According to Assumptions in Safety-Related Models for Automated Driving Systems
}

\begin{document}

\maketitle
\thispagestyle{empty}
\pagestyle{empty}
\copyrightnotice
\begin{abstract}
 In this work, we utilized the methodology outlined in the IEEE Standard 2846-2022 for ``Assumptions in Safety-Related Models for Automated Driving Systems” to extract information on the behavior of other road users in driving scenarios. This method includes defining high-level scenarios, determining kinematic characteristics, evaluating safety relevance, and making assumptions on reasonably predictable behaviors. The assumptions were expressed as kinematic bounds. The numerical values for these bounds were extracted using Python scripts to process realistic data from the UniD dataset. The resulting information enables Automated Driving Systems designers to specify the parameters and limits of a road user's state in a specific scenario. This information can be utilized to establish starting conditions for testing a vehicle that is equipped with an Automated Driving System in simulations or on actual roads. 
\end{abstract}

\section{Introduction}
\label{sec:introduction}

The development and deployment of \gls{ADS}-operated vehicles with level 3 automation or higher (according to the automation levels represented in the Society of Automotive Engineers (SAE) J3016 standard \cite{sae}, ranging from ``no driving automation" (level 0) to ``full driving automation" (level 5)) have the potential to improve safety compared to human drivers. However, it is clear that there will be still some level of risk associated with transportation even with the highest level of automation present.

Traffic rules are static regulations with the intention to guide road users safely through traffic \cite{Bogdoll2022}. Considering safety as the absence of unreasonable risk \cite{iso26262}, human road users generally follow these rules but not strictly. For instance, human drivers make assumptions about the behavior of other road users based on their daily experience, which helps them adjust their actions accordingly. In this sense, the interaction between human drivers, other road users (other vehicles, pedestrians, bicycles, etc), and the physical environment is not solely governed by traffic rules but by reasonable assumptions as well. The same reasoning applies to \gls{ADS}-operated vehicles, which will also need to make assumptions about the \textit{reasonably foreseeable} behavior of other road users\cite{iso26262}. In 2022 the IEEE Standard for Assumptions in Safety-Related Models for Automated Driving Systems \cite{ieee2846-2022} was published, defining the minimum set of assumptions that shall be considered for \gls{ADS}-operated vehicles and how they can be represented by bounds for kinematic variables. These assumptions enable the \gls{ADS}-operated vehicle to make informed decisions while considering the \textit{reasonably foreseeable} behavior of other road users and avoid unnecessary constraints. 

For the assumptions to qualify as \textit{reasonably foreseeable}, they need to be ``technically possible and with a credible or measurable rate of occurrence \cite{iso26262}". In this work, we extracted assumptions about the \textit{reasonably foreseeable} behavior of other road users (pedestrians, bicyclists, motorcyclists, and vehicles) using realistic data from the UniD dataset\cite{ind_dataset,round_dataset} in four common scenarios extracted from \cite{ieee2846-2022} and then tailored to the real environment where the data was recorded. 

The assumptions provided are intended to be used by \gls{ADS} designers when defining a safety envelope around the designed \gls{ADS}. This is a group of boundaries and requirements, within which the \gls{ADS} is designed to operate and to maintain operations within an acceptable risk level.

To the best of our knowledge, there have been no previous attempts in the literature to extract these assumptions from real data. 

The remainder of this paper is organized as follows: the next section describes related  work in  the  field. Section~\ref{sec:methodology} defines the assumptions and \ref{sec:use_cases} describes the scenarios considered in detail. Section~\ref{sec:implementation} presents an overview of the dataset and the method used to extract the relevant information. Section~\ref{sec:results} presents and discusses the obtained results. Finally, Section~\ref{sec:conclusion} concludes the present work outlining future research.

\section{Related Work}
\label{sec:relatedwork}

The automotive industry is already following functional safety requirements through standards, such as ISO 26262 \cite{iso26262}. However, operational safety requirements for ADS-Operated vehicles must also be provided in a standardized way to allow ADS designers to take relevant decisions in a well-informed manner.

In \cite{junietz2018}, the existing approaches for safety validation of ADS-operated vehicles are examined highlighting their limitations and simplifications. The authors identified four major approaches in the literature: Real world testing, Extreme Value Theory (EVT), scenario-based testing, and formal verification. All approaches have disadvantages and most of the time a combination of two or more is required to achieve a certain level of safety validation.
 
In \cite{Rodionova2020}, the authors acknowledged that making assurances for automated driving vehicles are a significant challenge in the industry. They present a method that utilizes the robustness of Metric Temporal Logic (MTL) specifications to explore the performance of safety models for automated vehicles. It includes a case study using the Responsibility Sensitive Safety model (RSS) and evaluates the trade-offs between safety and utility within the RSS parameter space, demonstrating that assertive driving behavior can achieve the highest utility while maintaining safety boundaries. The results were based on the CARLA driving simulator.

The \gls{NHTSA} started relevant efforts to standardize a framework to test safety scenarios in 2018 \cite{thorn2018}. They developed a preliminary testing framework based on three components: modeling and simulation, closed-track testing, and open-road testing. They also identified the importance of defining performance boundaries for safety. A similar methodology for generating test cases in the evaluation of \gls{ADS}' safety is proposed in \cite{thal2020}. The authors incorporate parameter correlations and safety relevance. By considering linear dependencies in safety-relevant situations and avoiding reliance on reference models, the proposed methodology addresses the limitations of existing approaches and contributes to ongoing discussions on standardized evaluation methodologies for automated driving safety.

In \cite{Neurohr2020} a simplified but more complete framework for scenario-based testing is introduced. The authors identify the arguments, principles, and assumptions behind each step of the scenario-based testing process and provide a list of fundamental considerations for which evidences need to be considered in order for scenario-based testing to support the homologation of ADS-operated vehicles. Similar considerations scenario-based approaches are presented in \cite{Lauer2022,Sippl2019,delre2022}.

In 2022, the IEEE Standard 2846-2022\cite{ieee2846-2022} was introduced. The Standard considers previous scenario-based frameworks and highlights the importance of assumptions in driving behavior for both human drivers and \gls{ADS}. Assumptions help human drivers establish a balance between conservativeness, safety, and natural driving, allowing for efficient traffic flow. Similarly, ADS-operated vehicles need to make assumptions about the reasonably foreseeable behaviors of other road users to inform their decision-making process. This standard aims to define these assumptions by setting bounds on kinematic variables to enable \gls{ADS} to make informed decisions based on what can be reasonably expected from other road users.

To contribute to the body of knowledge we present in this work the development of an approach that bases on kinematic characteristics to extract information from real data to make assumptions on reasonably predictable maneuvers according to the IEEE Standard 2846-2022\cite{ieee2846-2022}.

\section{Methodology}
\label{sec:methodology}
The methodology followed to derive the assumptions about \textit{reasonably foreseeable} behaviors of other road users was primarily based on the one presented in the standard \cite{ieee2846-2022} and is described as follows:
\begin{itemize}
    \item Definition of high-level scenarios describing particular driving situations including interactions with \gls{VRUs}. These scenarios are meant to be the building blocks for more complex scenarios. In some instances, a scenario may evolve from one to another. Each scenario definition should encompass the pertinent surroundings and include the identification and characterization of relevant road users.
    \item The pertinent kinematic characteristics that influence the movements of each road user in a given scenario, excluding the ego vehicle, were determined for the driving situation described in the scenario. For example, when a road user is in a lane adjacent to the ego vehicle, kinematics related to lateral motion become significant. For each kinematic property, its impact on safety was evaluated by determining if it has the potential to cause other road users to approach the ego vehicle. For example, if a road user next to the ego vehicle accelerates laterally, it will move toward the ego vehicle. Therefore, in the scenario under consideration, lateral acceleration is regarded as a kinematic attribute being safety-relevant. If the kinematic attribute is not safety-relevant, it is considered not applicable to the given scenario.
    \item Assumptions concerning \textit{reasonably foreseeable} behaviors of other road users were established for the given scenarios for each safety-relevant kinematic attribute of a given road user. The boundaries of what is considered reasonably predictable behavior for other road users in a given driving situation defined the assumptions.
\end{itemize}

As described in \cite{ieee2846-2022}, the assumptions take the form of constraints or bounding limits that specify the \textit{reasonably foreseeable} behavior of other road users:
\begin{itemize}
    \item $v^{lon}(t)\leq v^{lon}_{max}$: $v^{lon}(t)$ is the longitudinal velocity of other road users measured in $m/s$, and $v^{lon}_{max}$ is the \textit{reasonably foreseeable} maximum assumed value.
    \item $v^{lat}(t)\leq v^{lat}_{max}$: $v^{lat}(t)$ is the lateral velocity of other road users measured in $m/s$, and $v^{lat}_{max}$ is the \textit{reasonably foreseeable} maximum assumed value.
    \item $\alpha^{lon}(t)\leq \alpha^{lon}_{max}$: $\alpha^{lon}(t)$ is the magnitude of the longitudinal acceleration of other road users measured in $m/s^2$, and $\alpha^{lon}_{max}$ is the \textit{reasonably foreseeable} maximum assumed for that value.
    \item $\alpha^{lat}(t)\leq \alpha^{lat}_{max}$: $\alpha^{lat}(t)$ is the magnitude of the lateral acceleration of other road users measured in $m/s^2$, and $\alpha^{lat}_{max}$ is the \textit{reasonably foreseeable} maximum assumed for that value.
    \item $\beta^{lon}(t)\leq \beta^{lon}_{max}$: $\beta^{lon}(t)$ is the magnitude of the longitudinal deceleration of other road users measured in $m/s^2$, and $\beta^{lon}_{max}$ is the \textit{reasonably foreseeable} maximum assumed for that value.
    \item $\beta^{lon}(t)\geq \beta^{lon}_{min}$: $\beta^{lon}(t)$ is the magnitude of the longitudinal deceleration of other road users measured in $m/s^2$, and $\beta^{lon}_{min}$ is the \textit{reasonably foreseeable} minimum assumed for that value.
    \item $\beta^{lat}(t)\geq \beta^{lat}_{min}$: $\beta^{lat}(t)$ is the magnitude of the lateral deceleration of other road users measured in $m/s^2$, and $\beta^{lat}_{min}$ is the \textit{reasonably foreseeable} minimum assumed for that value.
    \item $h(t)\leq h_{max}$: $h(t)$ is the heading angle (yaw) of a road user measured in $^\circ$ (degrees), and $h_{max}$ is the \textit{reasonably foreseeable} maximum assumed for that value.
    \item $h'(t)\leq h'_{max}$: $h'(t)$ is the rate of change of the heading angle of a road user measured in $^\circ/s$, and $h'_{max}$ is the \textit{reasonably foreseeable} maximum assumed for that value.
    \item $\lambda(t)\leq\lambda_{max}$: $\lambda(t)$ is lateral fluctuation within a lane when the road user is moving forward measured in $m$, and $\lambda_{max}$ is the \textit{reasonably foreseeable} maximum assumed for that value.
\end{itemize}

The response time ($\rho$) of the road user is not included in this work due to the lack of related information in the selected dataset and the abundant literature regarding this specific parameter \cite{muttart2003}. It is important to notice that, in general, for each scenario only a subset of these general assumptions is applicable.

Once the assumptions and scenarios were  defined, the relevant cases were identified on the dataset and the numerical values for the assumption boundaries were extracted.

\section{Definition of Scenarios}
\label{sec:use_cases}

We decided to consider four different scenarios described in \cite{ieee2846-2022}  and then tailored to the data available in the UniD dataset.
\subsection{S1 - Ego vehicle driving next to other road users}
\label{sec:s1}
\begin{figure}[t]
	\centering
	\begin{subfigure}{0.237\textwidth}
		\includegraphics[width=\textwidth]{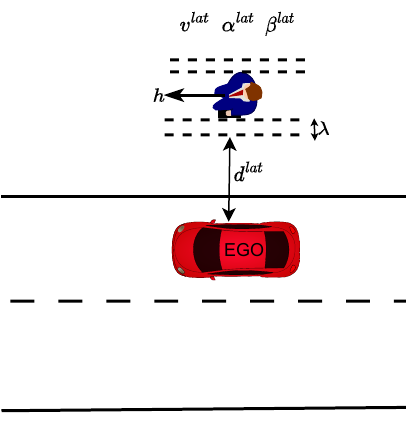}
		\caption{}
            \label{fig:example_s1a}
	\end{subfigure}
    \begin{subfigure}{0.237\textwidth}
		\includegraphics[width=\textwidth]{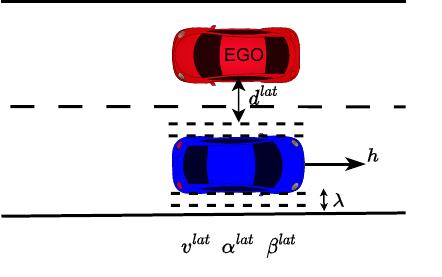}
		\caption{}
            \label{fig:example_s1b}
	\end{subfigure}
	\caption{Scenario S1: The ego vehicle (red) is driving next to another road user (a pedestrian in (a) or a blue car in (b)).}
	\label{fig:example_s1}
\end{figure}
In this scenario, the ego vehicle is driving alongside other road users who are traveling in the same direction. Both paths are expected to never intersect. A common example can be the ego vehicle driving next to a pedestrian on a sidewalk. Besides the ego vehicle, the other road users considered in this scenario can be:
\begin{itemize}
    \item Pedestrians: They can be found on the sidewalk and can either be moving in either direction relative to the ego vehicle or be stationary. Figure \ref{fig:example_s1a} provides an illustration of this scenario.
    \item  Cyclists: On the bike lane or on the road, they can be stationary or moving to the left or right of the ego vehicle.
    \item  Motorcyclist: On the road, they can be stopped or moving to the left or right of the ego vehicle.
    \item  Vehicles: On the road as well, they can be stationary or moving, laterally to the left of the ego vehicle and always driving in the opposite direction (due to limitations in the dataset). Figure \ref{fig:example_s1b} provides an illustration of this scenario.
\end{itemize}
Analyzing real data, it is possible to find situations when more than one road user is present at the same time (\textit{e.g.} groups of pedestrians walking together). In those cases, all the road users are considered separately as an independent instance of the same studied scenario.

The minimum set of assumptions for the road users in this case are:
\begin{itemize}
    \item $v^{lat}(t)\leq v^{lat}_{max}$
    \item $\alpha^{lat}(t)\leq \alpha^{lat}_{max}$
    \item $\beta^{lat}(t)\geq \beta^{lat}_{min}$
    \item $h(t)\leq h_{max}$
    \item $\lambda(t)\leq\lambda_{max}$
\end{itemize}

The first three are relevant for the lateral motion of the road user in the direction of the ego vehicle. $h_{max}$ is expected to be close to $0^\circ$ due to the scenario definition (parallel paths).

\subsection{S2 - Ego vehicle driving longitudinally behind another road user}
\label{sec:s2}
\begin{figure}[t]
    \centering
    \includegraphics[width=0.45\textwidth]{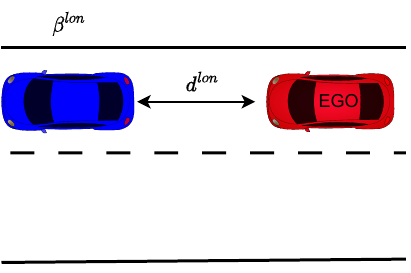}
    \caption{Scenario S2: The ego vehicle (red) is driving behind another road user (a blue car in this example).}
    \label{fig:example_s2}
\end{figure}

In this scenario, the ego vehicle is driving behind other road users who are traveling in the same direction. There are no road users behind the ego vehicle. Besides the ego vehicle, the other road user considered in this scenario can be:
\begin{itemize}
    \item Pedestrians: On the road, they are moving in the same direction as the ego vehicle and are located in front of it.
    \item  Cyclists: On the road, they are moving in the same direction as the ego vehicle and are located in front of it.
    \item  Motorcyclist: On the road, they are moving in the same direction as the ego vehicle and are located in front of it.
    \item  Vehicles: On the road, they are moving in the same direction as the ego vehicle and are located in front of it. An example of this case is depicted in Figure \ref{fig:example_s2}. 
\end{itemize}

The minimum set of assumptions for the road users in this case is:
\begin{itemize}
    \item $\beta^{lon}(t)\leq \beta^{lon}_{max}$
\end{itemize}

The longitudinal deceleration is the only safety-relevant kinematic property, for instance in the case of a sudden breaking maneuver.
\subsection{S3 - Ego vehicle driving in between leading and trailing road users}
\label{sec:s3}
\begin{figure}[t]
    \centering
    \includegraphics[width=0.45\textwidth]{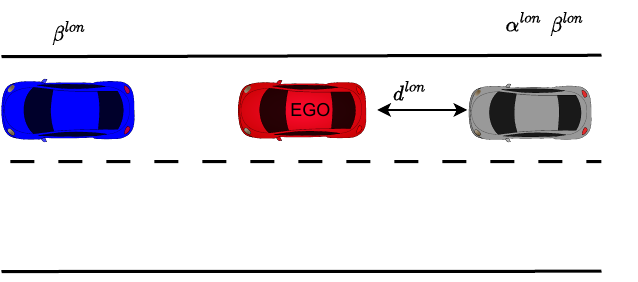}
    \caption{Scenario S3: The ego vehicle (red) is driving behind a road user (blue) and in front of another (grey). In this example both road users are cars.}
    \label{fig:example_s3}
\end{figure}

In this scenario, the ego vehicle is driving behind a leading road user and in front of a following road user, all traveling in the same direction on a road. Besides the ego vehicle, the other road users considered in this scenario can be:
\begin{itemize}
    \item Pedestrians: On the road, they are moving in the same direction as the ego vehicle and are located in front of it or/and behind it.
    \item  Cyclists: On the road, they are moving in the same direction as the ego vehicle and are located in front of it or/and behind it.
    \item  Motorcyclist: On the road, they are moving in the same direction as the ego vehicle and are located in front of it or/and behind it.
    \item  Vehicles: On the road, they are moving in the same direction as the ego vehicle and are located in front of it or/and behind it. An example of this case is depicted in Figure \ref{fig:example_s3}. 
\end{itemize}

The minimum set of assumptions for the trailing road user in this case is:
\begin{itemize}
    \item $\alpha^{lon}(t)\leq \alpha^{lon}_{max}$
    \item $\beta^{lon}(t)\geq \beta^{lon}_{min}$
\end{itemize}
The assumptions regarding the lateral movement of road users do not apply in this case and consequently, they are not considered a threat to safety due to the nature of this defined scenario. For the leading road user, the applicable set of assumptions was already defined in the previous scenario S2.
\subsection{S4 - VRU crossing the road in front of the Ego vehicle without crosswalk}
\label{sec:s4}
\begin{figure}[t]
    \centering
    \includegraphics[width=0.45\textwidth]{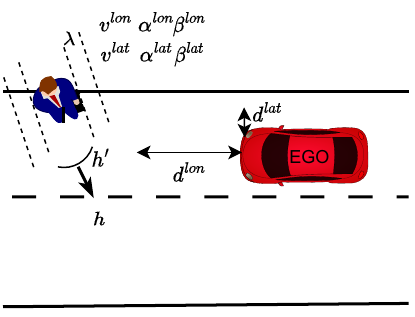}
    \caption{Scenario S4: The ego vehicle (red) is driving while a \gls{VRU} (blue pedestrian) is crossing the road.}
    \label{fig:example_s4}
\end{figure}

In this scenario, the ego vehicle is traveling along a road when a \gls{VRU} crosses or begins to cross the road without the use of a crosswalk or controlling signal.
\begin{itemize}
    \item Pedestrians: in motion or preparing to cross the street. An example of this case is depicted in Figure \ref{fig:example_s4}. 
    \item  Cyclists:  in motion or temporarily stopped and about to cross the road.
\end{itemize}

The minimum set of assumptions for this case is:
\begin{itemize}
    \item $v^{lon}(t)\leq v^{lon}_{max}$
    \item $v^{lat}(t)\leq v^{lat}_{max}$
    \item $\alpha^{lon}(t)\leq \alpha^{lon}_{max}$
    \item $\alpha^{lat}(t)\leq \alpha^{lat}_{max}$
    \item $\beta^{lon}(t)\leq \beta^{lon}_{max}$
    \item $\beta^{lon}(t)\geq \beta^{lon}_{min}$
    \item $\beta^{lat}(t)\geq \beta^{lat}_{min}$
    \item $h'(t)\leq h'_{max}$
    \item $\lambda(t)\leq\lambda_{max}$
\end{itemize}

In this case, both lateral and longitudinal motion of the \gls{VRU} can result in motion towards the vehicle, making them a relevant factor for safety considerations.

\section{Implementation}
\label{sec:implementation}
\subsection{Dataset}
\label{subsec:dataset}

To better understand naturalistic road users' behaviour, the UniD dataset was chosen  \cite{round_dataset,ind_dataset}. This dataset was recorded using a drone to overcome the occlusions of onboard sensors, equipped with a camera at a frame rate of 25 fps. The drone recorded trajectories of vehicles (cars, vans, busses), pedestrians, motorcyclists, and bicyclists. All recordings are fully annotated, including position, velocity, acceleration, heading, and class per road user. It was recorded on the RWTH Aachen University campus, where the speed limit is 50km/h (13.9m/s). Noteworthy is that the fraction of pedestrians to the total number of road users is significantly higher compared to the other datasets. A total number of 1124 vehicles and 7101 \glspl{VRUs} were captured in almost 3 hours of recording on weekdays.

\subsection{Extraction}
\label{subsec:extration}
To extract the numerical values for the \textit{reasonably foreseeable} assumptions, we coded a Python script to identify the defined scenarios within the dataset. The algorithm iteratively defines a car as the ego vehicle, looks for the defined scenarios for this car, and finally calculates all relevant variables of itself and other road users. These variables are stored in an accumulator. Afterwards, boundary values for each parameter can be defined using the maximum of each stored variable per scenario and class. These results are shown in Section \ref{sec:results}. 

\begin{figure}[t]
    \centering
    \includegraphics[width=0.45\textwidth]{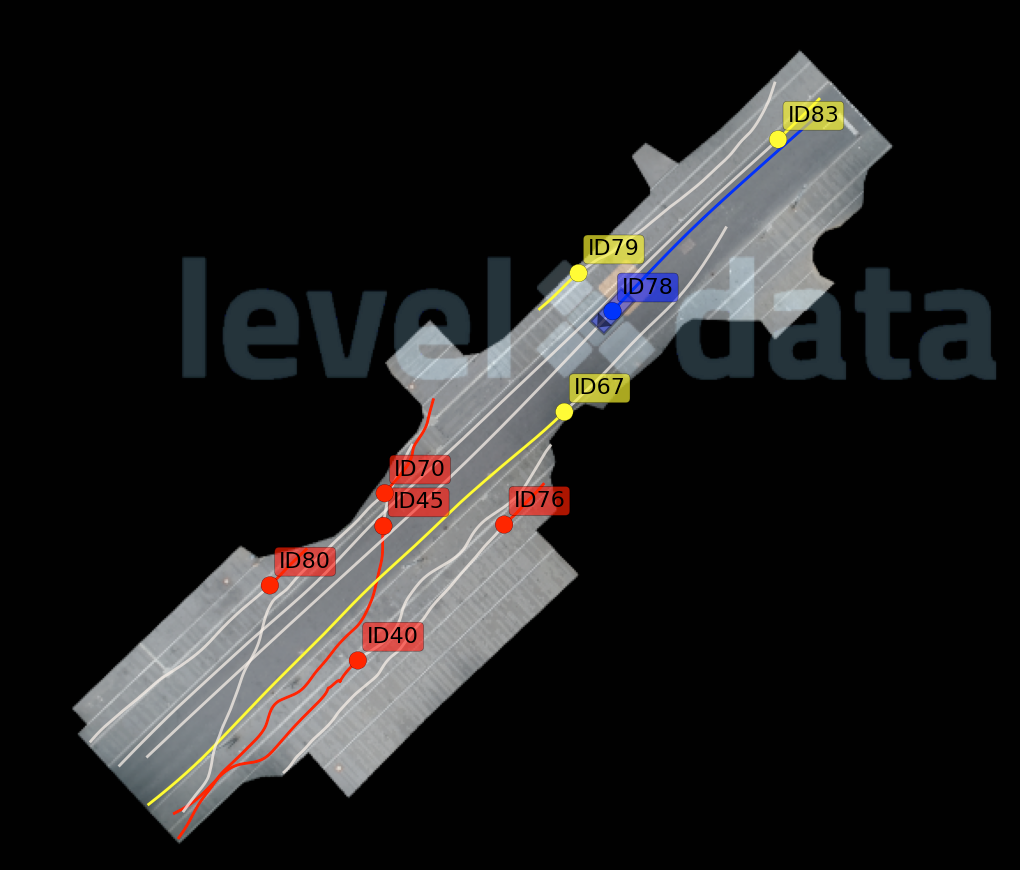}
    \caption{Selected time frame of the UniD dataset depicting Pedestrians (red), cyclists (yellow), and vehicles (blue). Considering ``ID78" as the ego vehicle, then there is one occurrence of scenario S1 with the cyclist ``ID79" and one occurrence of scenario S4 with the pedestrian ``ID45".}
    \label{fig:example1}
\end{figure}



It is possible for the same road user to participate in different scenarios at the same time or at different moments of the recording. Similarly, it is possible to have several occurrences or any given scenario overlapping in the same time frame. Figure \ref{fig:example1} shows a selected frame of the dataset where pedestrians are colored red, cyclists are colored yellow, and vehicles are colored blue. If we consider ``ID78" as the ego vehicle, there is one occurrence of scenario S1 with the cyclist ``ID79" and one occurrence of scenario S4 with the pedestrian ``ID45".


\section{Results}
\label{sec:results}
The resulting \textit{reasonably foreseeable} assumptions for each scenario are shown in Table \ref{tab:values_s1}, \ref{tab:values_s2}, \ref{tab:values_s3}, and \ref{tab:values_s4} including the number of times the use case ($N_{cases}$) was present in the studied dataset.

Considering scenario S1 (Table \ref{tab:values_s1}), we see that cars and pedestrians have much higher occurrences than Cyclists and Motorcyclists. Congruent with our intuition, pedestrians, and cyclists exhibited more sudden changes in heading ($h_{max}$) and had a higher lateral position fluctuation ($\lambda_{max}$). 

\begin{table}[ht]
\caption{Number of occurrences and boundary values for the assumptions of scenario S1}
\label{tab:values_s1}
\begin{center}
\begin{tabular}{lccccc}
Variable    & Units & Pedestrian & Cyclist & Motorcyclist & Vehicles  \\ \hline
$N_{cases}$ &  & 4416 & 73  & 92 & 2832 \\ 
 $v^{lat}_{max}$& $m/s$ & 0.2205 & 0.0  & 0.0 & 0.3119 \\ 
 $\alpha^{lat}_{max}$& $m/s^2$ & 0.5856 & 0.4202  & 0.8422 & 1.1822  \\ 
 $\beta^{lat}_{min}$& $m/s^2$ & 0.0001 &0.0003  & 0.0001 & 0.0001 \\ 
 $h_{max}$& $^\circ$ & 10.33  & 10.75  & 1.383 & 5.235 \\  
 $\lambda_{max}$& $m$ & 0.5545 & 0.6629 & 0.0245 & 0.1037\\ 
\end{tabular}
\end{center}
\end{table}

For scenario S2 (Table \ref{tab:values_s2}), there were only three occurrences involving pedestrians. The scenario required the pedestrian to walk longitudinally in front of the ego vehicle which was rare in this location due to the presence of a well-defined and wide sidewalk. Thus, the resulting bound might be not representative. For the rest of the road  users, the values are very similar and consistent with the described scenario, with the highest longitudinal deceleration for cyclists, which was about 22\% larger when compared to motorcyclists and vehicles. 

\begin{table}[ht]
\caption{Number of occurrences and boundary values for the assumptions of scenario S2}
\label{tab:values_s2}
\begin{center}
\begin{tabular}{lccccc}
Variable    & Units & Pedestrian & Cyclist & Motorcyclist & Vehicles   \\ \hline
 $N_{cases}$ &  & 3 & 132  & 17 & 328 \\ 
 $\beta^{lon}_{max}$& $m/s^2$ & 0.9543 & 2.4616 & 1.9365 & 2.0145 \\ 
\end{tabular}
\end{center}
\end{table}

Similar to scenario S2, the number of pedestrians in scenario S3 (Table \ref{tab:values_s3}) was too low to extract conclusive evidence. The bound for longitudinal acceleration
($\alpha^{lon}_{max}$) tends to grow according to the expected longitudinal force achievable for each road user. This means that the bound for longitudinal acceleration ($\alpha^{lon}_{max}$) is largest in vehicles, is lower for motorcyclists, and decreases more for cyclists, with pedestrians having the lowest ($\alpha^{lon}_{max}$) value. The deceleration bound is also consistent with the defined scenario.

\begin{table}[ht]
\caption{Number of occurrences and boundary values for the assumptions of scenario S3}
\label{tab:values_s3}
\begin{center}
\begin{tabular}{lccccc}
Variable    & Units & Pedestrian & Cyclist & Motorcyclist & Vehicles   \\ \hline
 $N_{cases}$ &  & 2 & 130  & 20 & 352 \\ 
 $\alpha^{lon}_{max}$& $m/s^2$ & 0.9038  &1.7879  & 2.1547 & 2.6553 \\ 
 $\beta^{lon}_{min}$& $m/s^2$ & 0.0042 & 0.0001  & 0.0001 & 0.0001 \\ 
\end{tabular}
\end{center}
\end{table}

For scenario S4 (Table \ref{tab:values_s4}), only occurrences involving pedestrians and cyclists were taken into account, according to the scenario definition. 
\begin{table}[ht]
\caption{Number of occurrences and boundary values for the assumptions of scenario S4}
\label{tab:values_s4}
\begin{center}
\begin{tabular}{lccc}
Variable    & Units & Pedestrian & Cyclist\\ \hline
 $N_{cases}$ &  & 1464 & 173 \\ 
 $v^{lon}_{max}$& $m/s$ & 8.24 & 13.13   \\ 
 $v^{lat}_{max}$& $m/s$ & 0.220 & 0.202  \\ 
 $\alpha^{lon}_{max}$& $m/s^2$ & 6.45 & 2.36   \\ 
 $\alpha^{lat}_{max}$& $m/s^2$ & 6.04 & 1.78  \\ 
 $\beta^{lon}_{max}$& $m/s^2$ & 2.96& 4.54   \\ 
 $\beta^{lon}_{min}$& $m/s^2$ & 0.0001 & 0.0001\\ 
 $\beta^{lat}_{min}$& $m/s^2$ & 0.0001 & 0.0001  \\ 
 $h'_{max}$& $^\circ/s$ & 4.40 & 2.933  \\  
 $\lambda_{max}$& $m$ &  6.174 & 4.53 \\ 
\end{tabular}
\end{center}
\end{table}

In general, the obtained results allow \gls{ADS} designers to parametrize and to bound the kinematic search space of a road user's state in a given scenario. Practically speaking, these results can be used to set up the scenario's initial conditions to test an \gls{ADS}-operated vehicle in simulation or on real roads. The decision-making process from the side of the \gls{ADS}-operated vehicle relies on two general inputs\cite{ieee2846-2022}:
\begin{itemize}
    \item the initial conditions of the road user which actually correspond to a measurement of the current state of the road users present in the environment at any given time.
    \item Boundaries on the reasonably anticipated actions that the road user(s) will take. 
\end{itemize}

By taking into account \textit{reasonably foreseeable} changes in the behavior of the nearby road users, these two inputs assist the \gls{ADS}-operated in planning its future actions. They can give the \gls{ADS}-operated vehicle the ability to weigh various options, such as choosing whether to accelerate and overtake a motorcyclist or widen its lateral gap. 

\section{Conclusion and Future Work}
\label{sec:conclusion}

This paper introduced a set of \textit{reasonably foreseeable} assumptions in the form of kinematic bounds for road users in four different scenarios by following the IEEE Standard for Assumptions in Safety-Related Models for Automated Driving Systems (IEEE Standard 2846-2022) \cite{ieee2846-2022}. The numerical values for these bounds were extracted from realistic data within a credible and measurable rate of occurrence.


Currently, the authors are working on the inclusion of more complex scenarios with occluded road users or limited visibility, intersections, roundabouts, and motorways. To this end, a comprehensive assessment of available datasets such as RounD\cite{round_dataset}, InD\cite{ind_dataset}, HighD\cite{highd_dataset}, and DORA \cite{dora_dataset} is being carried out, alongside the development of an entirely novel dataset from the vehicle perspective employing the JKU-ITS Vehicle for research \cite{its-vehicle}.

\section*{ACKNOWLEDGMENT}

This work was supported by the Austrian Science Fund (FWF), project number P 34485-N. It was additionally supported by the Austrian Ministry for Climate Action, Environment, Energy, Mobility, Innovation, and Technology (BMK) Endowed Professorship for Sustainable Transport Logistics 4.0., IAV France S.A.S.U., IAV GmbH, Austrian Post AG and the UAS Technikum Wien.

\bibliographystyle{IEEEtran}
\bibliography{paper}
\end{document}